\documentclass[sigconf]{acmart}

\usepackage{multirow}

\usepackage{algorithm}
\usepackage[noend]{algpseudocode}

\algrenewcommand\alglinenumber[1]{#1}

\usepackage{amsmath,amssymb,amsfonts}
\usepackage{graphicx}
\usepackage{textcomp}
\usepackage{xcolor}
\usepackage{bbding}
\usepackage{epstopdf}
\usepackage{epsfig}
\usepackage{enumitem}
\setlist[itemize]{leftmargin=*, topsep=0.2em, itemsep=0pt, parsep=0pt, partopsep=0pt}
\setlist[enumerate]{leftmargin=*, topsep=0.2em, itemsep=0pt, parsep=0pt, partopsep=0pt}

\newlength{\enumerateparindent}
\usepackage{bbm}

\usepackage{collectbox}

\makeatletter
\newcommand{\mybox}{%
    \collectbox{%
        \setlength{\fboxsep}{1pt}%
        \fbox{\BOXCONTENT}%
    }%
}
\makeatother

\copyrightyear{2024}
\acmYear{2024}
\setcopyright{rightsretained}
\acmConference[CIKM '24]{Proceedings of the 33rd ACM International Conference on Information and Knowledge Management}{October 21--25, 2024}{Boise, ID, USA}
\acmBooktitle{Proceedings of the 33rd ACM International Conference on Information and Knowledge Management (CIKM '24), October 21--25, 2024, Boise, ID, USA}
\acmDOI{10.1145/3627673.3679870}
\acmISBN{979-8-4007-0436-9/24/10}
\makeatletter
\gdef\@copyrightpermission{
}  
\makeatother

\begin{document}
\title{A Systematic Evaluation of Generated Time Series and\\Their Effects in Self-Supervised Pretraining}

\author{Audrey Der$^\dag$, Chin-Chia Michael Yeh*, Xin Dai*, Huiyuan Chen*, Yan Zheng*, Yujie Fan*,\\Zhongfang Zhuang*, Vivian Lai*, Junpeng Wang*, Liang Wang*, Wei Zhang*, Eamonn Keogh$^\dag$ \\ ader003@ucr.edu, miyeh@visa.com}
\affiliation{%
    \institution{Visa Research*; University of California, Riverside$^\dag$}
    \state{Foster City, California; Riverside, California}
    \country{USA}
}

\renewcommand{\shortauthors}{Audrey Der et al.} 

\begin{abstract}
Self-supervised Pretrained Models (PTMs) have demonstrated remarkable performance in computer vision and natural language processing tasks.
These successes have prompted researchers to design PTMs for time series data. 
In our experiments, most self-supervised time series PTMs were surpassed by simple supervised models.
We hypothesize this undesired phenomenon may be caused by data scarcity.
In response, we test six time series generation methods, use the generated data in pretraining in lieu of the real data, and examine the effects on classification performance.
Our results indicate that replacing a real-data pretraining set with a greater volume of only generated samples produces noticeable improvement.
\end{abstract}


\maketitle

\sloppy
Self-supervised Pretrained Models (PTMs) are models initially trained using self-supervised losses~\cite{ma2023survey}, not requiring labeled samples.
The unsupervised nature of this process allows PTMs to utilize a substantial amount of unlabeled data, and more robust features are extracted for intended tasks.
The success of PTMs in computer vision (CV)~\cite{chen2020simple} and natural language processing (NLP)~\cite{qiu2020pre} tasks motivate the development of PTMs for time series~\cite{ma2023survey}.

\begin{figure}[ht]
\centerline{
\includegraphics[width=0.99\linewidth]{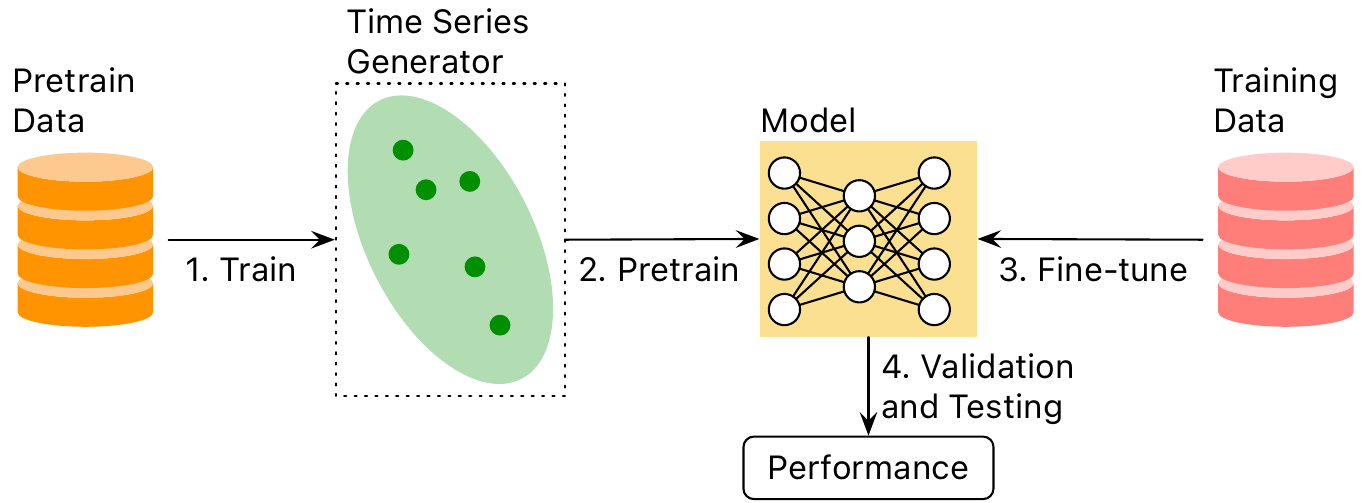}
}
\caption{
The time series generator can be utilized to synthesize data for self-supervised pretraining.
}
\label{fig:problem}
\vspace{-1em}
\end{figure}

We integrate four self-supervised PTMs (TS2Vec~\cite{yue2022ts2vec}, MixingUp~\cite{wickstrom2022mixing}, TF-C~\cite{zhang2022self}, TimeCLR~\cite{yeh2023toward}) with two popular network architectures (ResNet~\cite{he2016deep,wang2017time} and Transformer~\cite{vaswani2017attention}), but found that pretraining does not always enhance the performance of the model for time series classification (TSC). 
Figure~\ref{fig:problem} describes our problem setting.
Rather than pretraining with a pretraining split, we use that data to derive a time series generator, which is used to synthesize a large volume of data for pretraining in its stead.
Once pretrained, we apply standard supervised training to fine-tune the model using training data.
Then, the model is validated and tested using the validation and test splits.

This work undertakes a comparative analysis to understand if and how using different generated time series during pretraining affects TSC performance. 
We experiment with combinations of one of four PTMs, one of six time series generators, and one of two network architectures. 
Our findings suggest using generated time series in pretraining enhances the performance of PTMs compared to using an equally-sized or smaller pretraining set consisting of real data.
\section{Related Work}
We conduct a comparative study on the interplay between time series generation, pretraining, and classification. 
We discuss related work from each of these areas.

\noindent \textbf{Generation.} 
Time series generation has been a popular research area due to the success of generative models in computer vision and natural language processing~\cite{wen2020time}.
For example, Generative Adversarial Networks (GANs) have been used to generate time series for several domains, including music, speech, and EEG signals~\cite{zhang2022comprehensive}.
Variational Autoencoders (VAEs)~\cite{kingma2013auto} have been used for anomaly detection~\cite{guo2018multidimensional, matias2021robust}, multivariate adversarial time series generation~\cite{harford2021generating}, and imputation~\cite{li2021variational}.
Diffusion models~\cite{ho2020denoising} have exploded in popularity over the last few years, with practitioners applying diffusion models to time series~\cite{lin2023diffusion} for forecasting~\cite{rasul2021autoregressive,alcaraz2023diffusionbased}, time series imputation~\cite{tashiro2021csdi,alcaraz2023diffusionbased}, and generating waveforms~\cite{li2019enhancing,chen2020wavegrad}.
Data generation in time series classification has proven beneficial, of which GANs, VAEs, and diffusion models have been leveraged~\cite{iwana2020time,wen2020time}. 
The success of these generative models motivates us to examine outcomes using generated time series in pretraining.

\noindent \textbf{Pretraining.}
Research on time series PTMs is a burgeoning area of study in the field~\cite{ma2023survey}. 
This task, however, needs to be handled with care because of several complexities. 
These include inconsistent or unknown domain invariances, the wide semantic meaning across data types, varying sampling rates, and differing data sources~\cite{zhang2022self,gupta2020transfer, meiseles2020source}. 
Despite these complexities, these methods have shown promising results in a foundation model setting~\cite{yeh2023toward}, where multiple datasets are used for pretraining. 
Taking inspiration from~\cite{yeh2023toward}, we have conducted experiments using four self-supervised contrastive learning pretraining methods: MixingUp~\cite{wickstrom2022mixing}, TimeCLR~\cite{yeh2023toward}, TF-C~\cite{zhang2022self}, and TS2Vec~\cite{yue2022ts2vec}.


\noindent \textbf{Classification.} 
There is an extensive body of work on time series classification, with ensemble and tree-based methods in HIVE-COTE 2.0~\cite{middlehurst2021hive} and TS-CHIEF~\cite{shifaz2020ts} being state-of-the-art  until recently. 
Another family of methods that has achieved top-tier performance include ROCKET and its variants~\cite{tan2022multirocket}, which extract features using convolutional kernels.
High-performing deep learning models such as FCN~\cite{wang2017time}, ResNet~\cite{wang2017time}, and InceptionTime~\cite{ismail2020inceptiontime} all utilize convolution-based designs.
Notably, Fawaz et al.~\cite{ismail2019deep} have demonstrated ResNet~\cite{wang2017time} to be a competitive baseline in time series classification.
We elect to include ResNet~\cite{wang2017time} as the representative convolution-based architecture design in our experiments.
Transformers have been another popular choice of architecture for time series data~\cite{ahmed2023transformers,yuan2020self,wen2022transformers}. 
Good classification performance from methods such as AutoTransformer~\cite{ren2022autotransformer} and Gated Transformer Network~\cite{liu2021gated}, motivate including the Transformer into our time series classification experiments.

To the best of our knowledge, no other systematic evaluation has assessed the interaction between the time series generation method and the pretraining method.

\section{Model Architecture}
\label{sec:model}
We focus on two widely utilized model architectures for time series data: the \textit{ResNet} and \textit{Transformer}.

\noindent \textit{ResNet}: 
The Residual Network (ResNet) is a time series classification model that takes inspiration from the success of ResNet as it was first introduced in computer vision~\cite{he2016deep,wang2017time}. 
Extensive evaluation by~\cite{ismail2019deep} have demonstrated ResNet as one of the strongest models for time series classification.

\begin{figure}[ht]
\centerline{
    \includegraphics[width=0.85\linewidth]{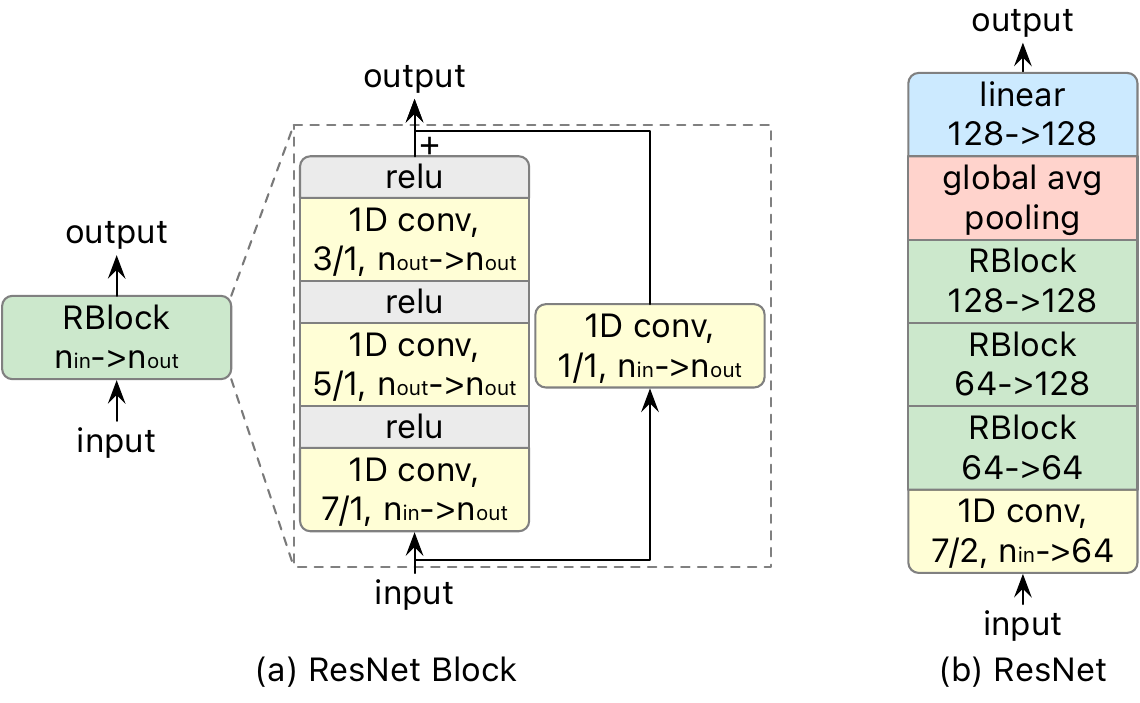}
    }
    \caption{The designs of the residual block, \protect\mybox{\texttt{RBlock}}, and the ResNet, are shown in this figure.}
    \label{fig:resnet}
\end{figure}

The specific design we use is shown in Figure~\ref{fig:resnet}.b, and is based on the design proposed by~\cite{wang2017time}. 
We use the notation \mybox{\texttt{RBlock, 64$\to$64}} to denote a residual block (Figure~\ref{fig:resnet}.a) with an input dimension of 64 and an output dimension of 64.

The design of the residual block can be found in Figure~\ref{fig:resnet}.a. 
This block consists of the main passage and the skip connection passage. 
The main passage processes the input sequence using three pairs of $1D$ convolutional-ReLU layers. 
The convolutional layers have filter sizes of seven, five, and three, sequentially, progressing from the input to the output.
On the other hand, the skip connection passage may include an optional $1D$ convolutional layer with a filter size of one. 
This convolutional layer is only introduced to the skip connection when the input dimension and the output dimension of the residual block differ.
The output of the main passage and the skip connection passage are combined through element-wise addition to form the final output of the block. 

\begin{figure}[ht]
\centerline{
\includegraphics[width=0.85\linewidth]{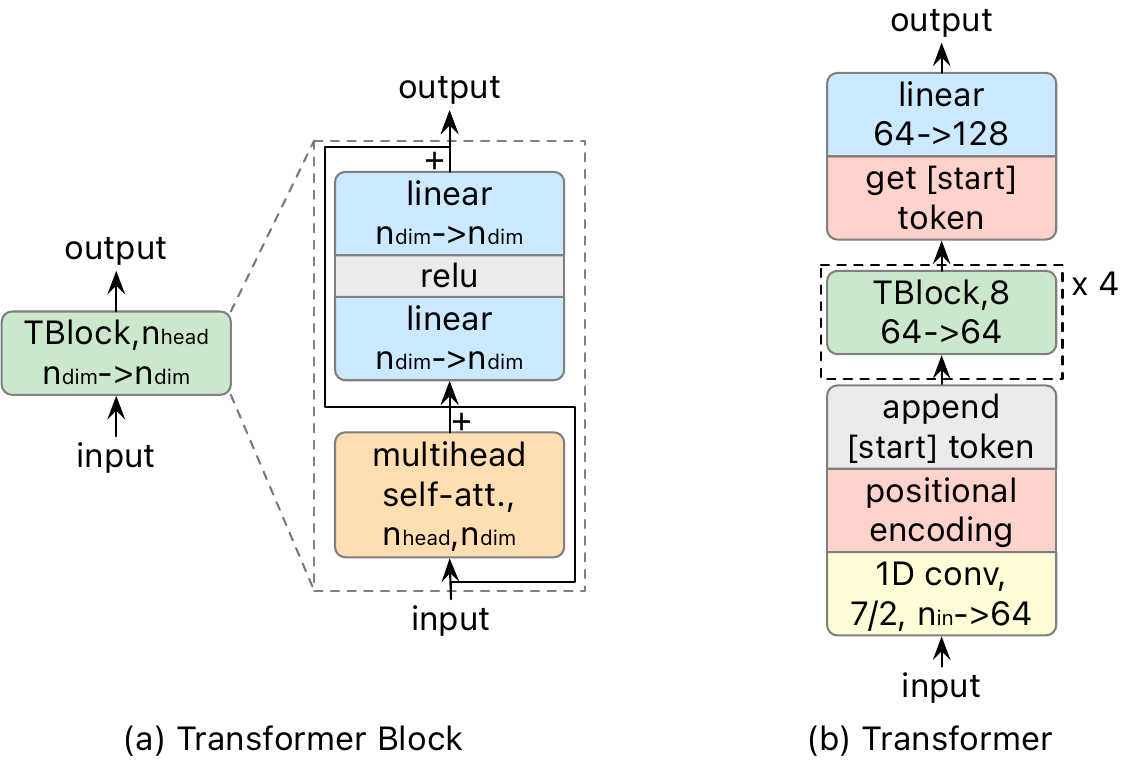}
}
\caption{
The designs of the transformer block, \protect\mybox{\texttt{TBlock}}, and the Transformer, are shown in this figure.
}
\label{fig:transform}
\end{figure}

\noindent \textit{Transformer}: 
The Transformer is a widely used architecture for sequence modeling~\cite{vaswani2017attention,li2019enhancing,zhou2021informer,lim2021time,chen2022denoising}. 
The architecture we used is shown in Figure~\ref{fig:transform}.b. 
We used fixed positional encoding (following~\cite{vaswani2017attention}) and included a special token, \texttt{[start]}, to learn the representation of the entire time series. 
Our Transformer architecture consists of four encoder layers, each denoted as \mybox{\texttt{TBlock,8,64${\to}$64}} (Figure~\ref{fig:transform}.a) where the number of heads is 8, the input dimension is 64, and the output dimension is 64.

The Transformer block (illustrated in Figure~\ref{fig:transform}.a) is composed of a multihead self-attention stage and a feed-forward stage.
The multihead attention module allows the Transformer block to capture dependencies between different positions in the input sequence.
The position-wise feed-forward network applies two linear layers with a ReLU activation function to each position of the sequence obtained from the multihead self-attention module.
This enables the Transformer block to incorporate non-linear transformations and enhance the representation of each position.
Both stages incorporate skip connections, ensuring that the input is added to the output at each stage. 
This mechanism facilitates the flow of information from the input to the output, enabling the model to retain important information throughout the block.
The Transformer block is responsible for modeling complex dependencies and enhancing the representation of the input sequence through self-attention and position-wise transformations.

We use layer normalization~\cite{ba2016layer} for all normalization layers, as it is both effective and widely used in modeling sequential data~\cite{ba2016layer,vaswani2017attention}.
\section{Pretraining Methods}
\label{sec:pretrain}
Our experiments considered four different pretraining methods, all of which are based on constrastive learning.

\begin{figure}[ht]
\centerline{
\includegraphics[width=0.99\linewidth]{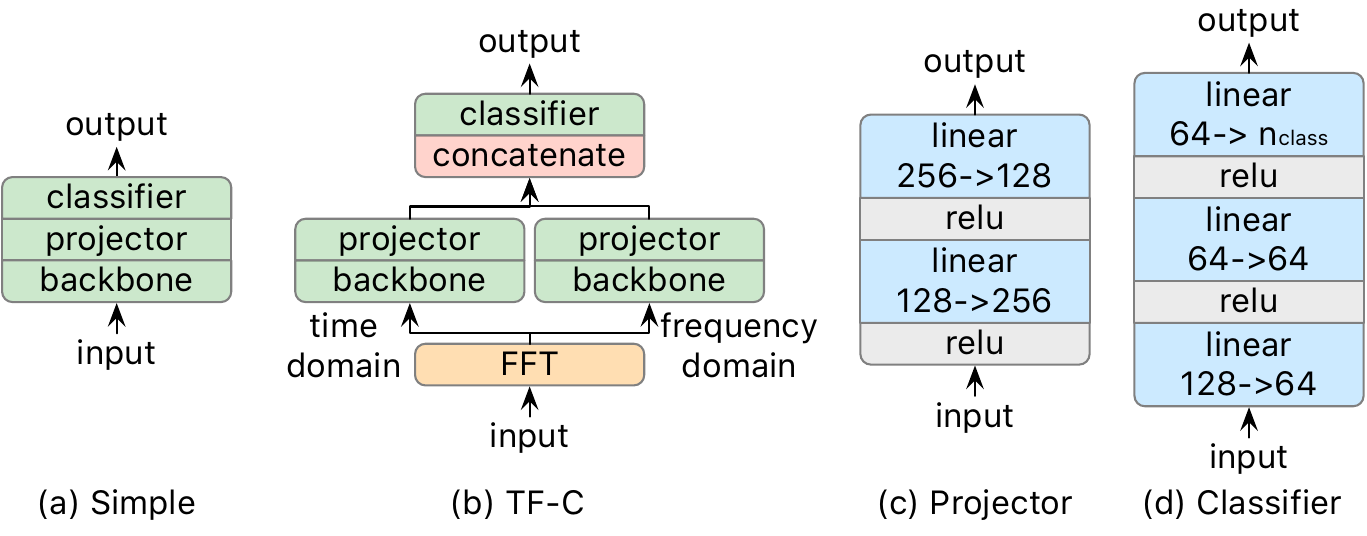}
}
\caption{
There are two ways to use backbone models for pretraining and finetuning.
All pretraining methods but TF-C use Figure~\ref{fig:pretrain}.a. TF-C uses Figure~\ref{fig:pretrain}.b.
The projector and classifier are shown in Figures~\ref{fig:pretrain}.c and~\ref{fig:pretrain}.d.
}
\label{fig:pretrain}
\end{figure}

\noindent \textit{TimeCLR}:
TimeCLR is based on SimCLR~\cite{tang2020exploring, yeh2023toward}, which was originally proposed as a self-supervised pretraining method based on contrastive learning for computer vision~\cite{chen2020simple}, and was later extended to human activity time series by~\cite{tang2020exploring}.
During pretraining, random scaling and negation augmentations~\cite{zhang2022self, contributions2023tfcgithub} are randomly applied to a batch of time series $X$ to generate two augmented batches $X_0$ and $X_1$.
Both $X_0$ and $X_1$ are then processed by the backbone model (Section~\ref{sec:model}) and the projector (Figure~\ref{fig:pretrain}.c). The output feature vectors are denoted as $H_0$ and $H_1$, respectively.

\sloppy
The NT-Xent loss function~\cite{chen2020simple,tang2020exploring} is used to compute the loss. 
If $h_i \in H_0$ and $h_j \in H_1$ are features extracted from the augmented versions of the same  series in $X$, the loss for the positive pair $(h_i, h_j)$ is computed as follows:
\begin{equation*}
\scriptsize
\mathcal{L} = - \log \frac{\text{exp} \left(\text{sim}(h_i, h_j)/\tau\right)}{\sum_{h_k \in H_0 + H_1} \mathbbm{1}_{[h_k \neq h_i \& h_k \neq h_j]} \text{exp}\left(\text{sim}(h_i, h_k)/\tau\right)}
\end{equation*}
Here, $\text{sim}(\cdot, \cdot)$ computes the cosine similarity between the input vectors, where $h_k$ is a feature vector from $H_0$ or $H_1$ that is not $h_i$ nor $h_j$, and $\tau$ is the temperature parameter.
Both the backbone model and projector are optimized using the NT-Xent loss. 
We add a classifier (Figure~\ref{fig:pretrain}.d) on top of the projector, as shown in Figure~\ref{fig:pretrain}.a, to fine-tune the model for the classification task. The backbone, projector, and classifier are updated using the cross-entropy loss.

TimeCLR improves SimCLR by incorporating additional augmentation techniques based on recent survey papers~\cite{wen2020time,iwana2021empirical,yeh2023toward}.
These augmentations are jittering, smoothing, magnitude warping, time warping, circular shifting, adding slope, adding spike, adding step, masking, and cropping~\cite{yeh2023toward}. 
These additional techniques enhance the pretraining process by allowing the model to learn the augmentations as potential invariances.
Jittering and smoothing aid the model's robustness to Gaussian noise, and adding warping to magnitude and time warping help the model become invariant. 
Circular shifting, adding slopes, spikes, steps, and masking make the model robust to corresponding noise types, while cropping helps the model learn contextual consistency similar to TS2Vec~\cite{yue2022ts2vec}.

TimeCLR only uses \textit{one} augmentation function for generating positive pairs instead of using all augmentation functions~\cite{zhang2022self, contributions2023tfcgithub}. 
Applying all augmentation methods simultaneously to a time series may produce augmented time series bearing little visual similarity to the original. 
Applying all the augmentation methods to a simple pattern (\includegraphics[height=0.9em]{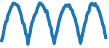}) results in a series (\includegraphics[height=0.9em]{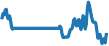}), which has lost many of the properties of the original time series.
For fine-tuning, we add a classifier model on top of the projector (as shown in Figure~\ref{fig:pretrain}.a) and update the backbone, projector, and classifier model using the cross-entropy loss. 


\noindent \textit{TS2Vec}:
The TS2Vec~\cite{yue2022ts2vec} generates positive pairs using the contextual consistency strategy, which generates positive pairs by randomly cropping two overlapping subsequences from a time series. 
The TS2Vec method computes the contrastive loss in a heirarchial fashion under multiple levels of time granularity. 
At each time granularity, two contrastive losses are computed: temporal contrastive loss and instance-wise contrastive loss. 
Both are modified NT-Xent losses where the former helps the model learn discriminative representations over time, and the latter helps the model learn discriminative representations over samples.

\sloppy
TS2Vec requires the backbone model to generate representation for multiple time steps. 
To achieve this, we modify the ResNet (Figure~\ref{fig:resnet}) by removing the \mybox{\texttt{global avg pooling}} block and apply the subsequent linear layer and projector model independently at each time step to generate the per-time-step representation. 
Likewise, we remove the \mybox{\texttt{get [start] token}} block from the Transformer (Figure~\ref{fig:transform}), and apply the subsequent linear layer and projector model independently at each time step to generate the per-time-step representation. 
The first layer for the backbone models is a $1D$ convolutional layer with a stride size of 2, and the length of the output sequence will be half that of the input time series. 
To fine-tune the model for classification, we re-add the removed blocks back to the model and add a classifier model (Figure~\ref{fig:pretrain}.d) on top of the projector; this is  shown in Figure~\ref{fig:pretrain}.a. Then the backbone, projector, and classifier are updated using the cross-entropy loss.

\noindent \textit{MixingUp}:
Given a pair of time series $(x_i, x_j)$, the MixingUp method~\cite{wickstrom2022mixing} generates a mixed time series $x_k$ by computing $\lambda x_i + (1-\lambda) x_j$. 
The mixing parameter $\lambda$ is randomly drawn from a beta distribution and determines the contribution of $x_i$ and $x_j$ in $x_k$.
The pretraining self-supervised task is predicting $\lambda$ from the representations associated with $x_i$, $x_j$, and $x_k$. 
The representations are generated by processing $x_i$, $x_j$, and $x_k$ using the backbone and projector model. 
The loss function adopted by~\cite{wickstrom2022mixing} is a modified version of the NT-Xent loss.
We handle fine-tuning similarly to the other pretraining models; a classifier is added to the projector (Figure~\ref{fig:pretrain}.a), and updating the backbone, projector, and classifier model is done using the cross-entropy loss.

\noindent \textit{TF-C}:
The TF-C method extends the idea of contrastive learning to the frequency domain~\cite{zhang2022self}. 
Based on the open-source implementation~\cite{zhang2022self,contributions2023tfcgithub}, the self-supervised learning procedure is as follows:
the method transforms a time series from the time domain to the frequency domain using the Fast Fourier transform (FFT). 
Positive pairs are generated through augmentation functions applied to the time series in both the time and frequency domains. 
TF-C uses jittering to augment the time series in the time domain and adds and removes frequency components in the frequency domain. 

The input time series are processed in both the time and frequency domains using their respective backbone models and projectors.
The self-supervised learning loss is calculated using a modified NT-Xent loss function, which involves intermediate representations from both the time and frequency domains.
Because the TF-C method uses two backbone models and two projector models, the width of these models is halved so the  number of parameters will be equal to that of the other methods. 
During fine-tuning, the output of the projector in time and frequency domain is concatenated before being fed into the classifier (Figure~\ref{fig:pretrain}.b), and is performed using cross-entropy loss.

\section{Generative Model}
In this paper, we examine three simple time series generators and three generative models. 

\begin{figure}[ht]
\centerline{
\includegraphics[width=0.8\linewidth]{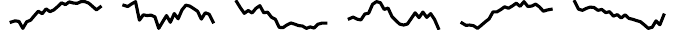}
}
\caption{
Generated random walk signals.
}
\label{fig:gen_ex_rw}
\end{figure}

\noindent \textit{Random Walk (RW)}: 
This generator produces random walk time series of specified length and dimensionality. 
Pretraining with random walk may be beneficial due to its ability to simulate various real-world time series, including fluctuating stock prices, Brownian motion, and the unpredictable paths of animals searching for food in the wild.
Figure~\ref{fig:gen_ex_rw} shows time series generated by our random walk generator.

\begin{figure}[ht]
\centerline{
\includegraphics[width=0.8\linewidth]{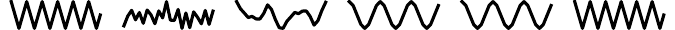}
}
\caption{
Generated sinusoidal waves.
}
\label{fig:gen_ex_sw}
\end{figure}

\noindent \textit{Sinusoidal Wave (SW)}: 
For each dimension, this method generates two sinusoidal time series of a specified length, utilizing randomly sampled frequencies, amplitudes, and offsets. 
These generated time series are combined to form each dimension of the output time series. 
Sinusoidal waves may serve as an ideal stand-in for real periodic time series.
Figure~\ref{fig:gen_ex_sw} shows time series samples from our sinusoidal wave generator.

\begin{figure}[ht]
\centerline{
\includegraphics[width=0.95\linewidth]{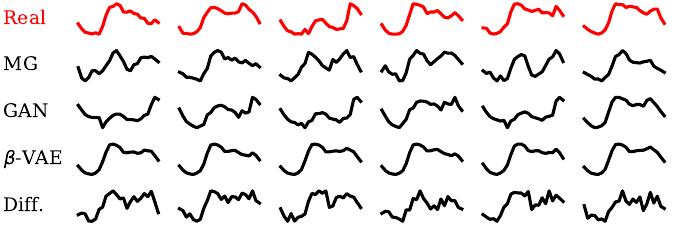}
}
\caption{
Real time series data from the UCR Archive \textit{ItalyPowerDemand} dataset are pictured in the first row, colored \textcolor{red}{red}. Below them are time series generated by the MG generator, GAN, $\beta$-VAE, and diffusion model.
}
\label{fig:gen_ex}
\end{figure}

\noindent For the next four methods (MG, GAN, $\beta$-VAE, and Diff.), we contrast six time series from each method with six samples of real data in Figure~\ref{fig:gen_ex}. 

\noindent \textit{Multivariate Gaussian (MG)}: 
We model the dataset as a simple multivariate Gaussian distribution in frequency domain, computing the mean and variance for each frequency from the pretraining set.
Time series are generated by sampling this distribution.
We do not consider the covariance between different frequencies.

\noindent \textit{Generative Adversarial Network (GAN)}: 
We develop a training framework for the GAN, taking inspiration from the BiGAN~\cite{donahue2016adversarial}. 
We have one major difference in our implementation; given our emphasis on time series data, we utilize a 1D (rather than 2D) convolutional network design tailored to this type of data.

\begin{figure}[ht]
\centering
\includegraphics[width=0.85\linewidth]{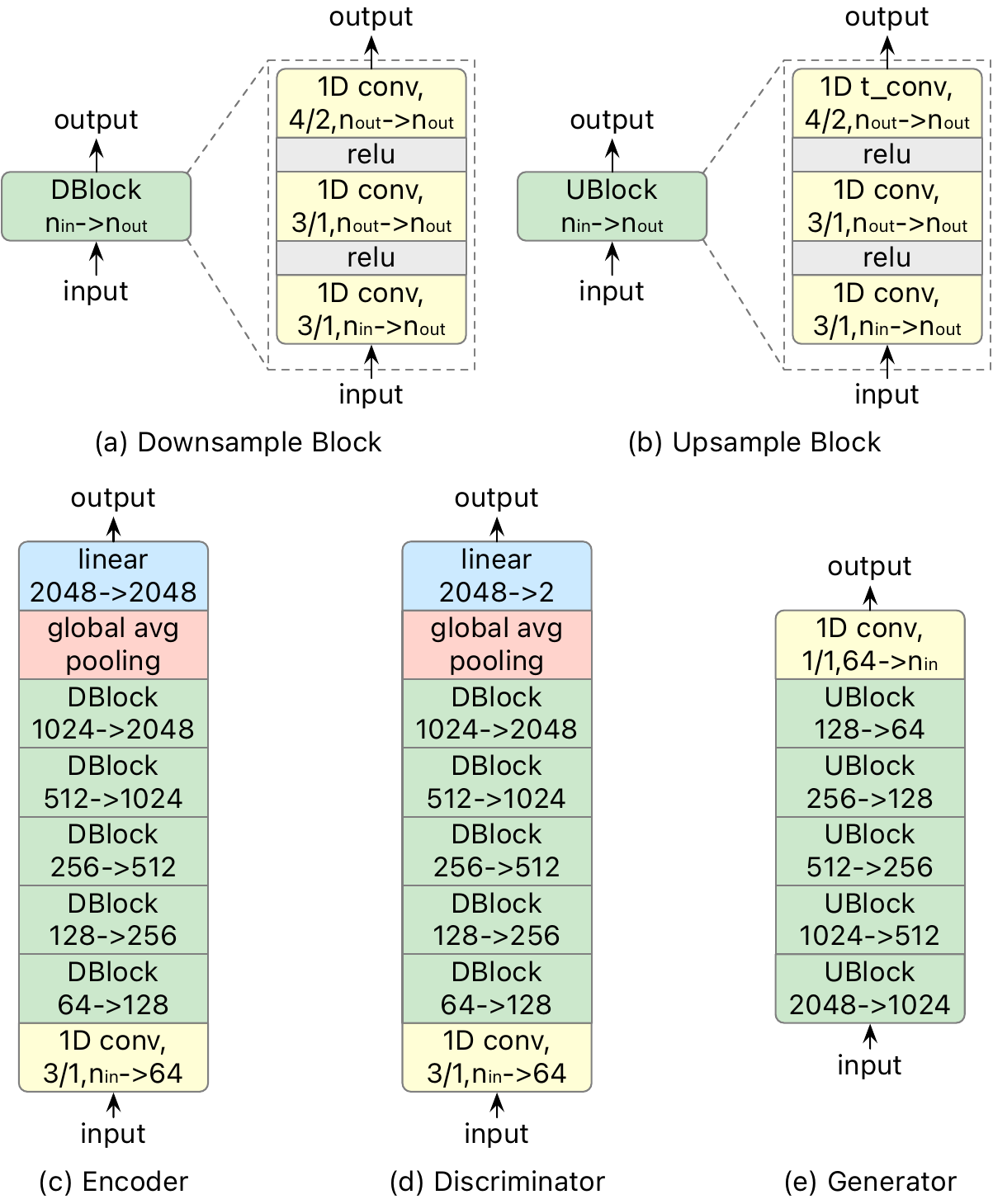}
\caption{The designs of the encoder, discriminator, and generator for our GAN implementation.}
\label{fig:gan_design}
\end{figure}

In our implementation, we utilize three sub-networks (shown in Figure~\ref{fig:gan_design}): the encoder, generator, and discriminator. 
The downsample block \mybox{\texttt{DBlock, $n_\text{in}\to n_\text{out}$}} reduces the temporal resolution of the input sequence, while the upsample block \mybox{\texttt{UBlock, $n_\text{in}\to n_\text{out}$}} increases the temporal resolution of the input sequence. 
In each iteration of the training process, we train the encoder and generator with mean squared error reconstruction loss: 
\begin{equation*}
\mathcal{L}_\text{reconstruction} = \text{MSE}(X, G(E(X)))
\end{equation*}
where $X$ denotes the training samples, $G(\cdot)$ denotes the generator and $E(\cdot)$ the encoder. 
The discriminator is trained with a critic loss~\cite{arjovsky2017wasserstein}: 
\begin{equation*}
\mathcal{L}_\text{critic} = \text{mean}(D((G(Z)) - \text{mean}(D(X)))
\end{equation*}
where $Z$ denotes random vectors and $D(\cdot)$ the discriminator. 
A generator loss~\cite{arjovsky2017wasserstein} is used to train $G(\cdot)$: 
\begin{equation*}
\mathcal{L}_\text{generator} = -\text{mean}(D(G(Z)))
\end{equation*}
We employ the same critic loss and generator loss as those used in the Wasserstein GAN. ~\cite{arjovsky2017wasserstein}.

\noindent \textit{$\beta$-Variational Autoencoder ($\beta$-VAE)}: 
The $\beta$-Variational Autoencoder (VAE)~\cite{higgins2016beta} is another effective approach to model data distribution for data generation. 
This method utilizes the reparameterization trick, which aids the VAE in learning the data distribution within the latent space~\cite{kingma2013auto}. 
Instead of generating a fixed latent representation for each input sample, the VAE generates parameters linked to a specific distribution (usually a Gaussian distribution). 
The $\beta$-VAE further improves the learning potential of the VAE by introducing a hyper-parameter, $\beta$, which controls the relative importance of different loss terms.

The $\beta$-VAE is comprised of an encoder and a decoder. 
Our encoder design is similar to the $1D$ convolutional design in our GAN implementation (i.e., Figure~\ref{fig:gan_design}.c).
In this figure, there is only a single output \mybox{\texttt{Linear,2048${\to}$2048}} layer. 
However, the encoder in our $\beta$-VAE differs in the design of the output layer as it needs to generate both the mean~$\mu$ and the log variance $\log{\sigma^2}$ for the Gaussian distribution. 
Consequently, our $\beta$-VAE encoder has two parallel output \mybox{\texttt{Linear,2048${\to}$2048}} layers\footnote{One layer generates $\mu$, while the other produces $\log{\sigma^2}$.}, in contrast to the single output layer depicted in Figure~\ref{fig:gan_design}.c.
The decoder is depicted in Figure~\ref{fig:gan_design}.e. 
We use the standard $\beta$-VAE training procedure~\cite{higgins2016beta} to train the encoder and decoder.



\begin{figure}[ht]
\centerline{
\includegraphics[width=0.5\linewidth]{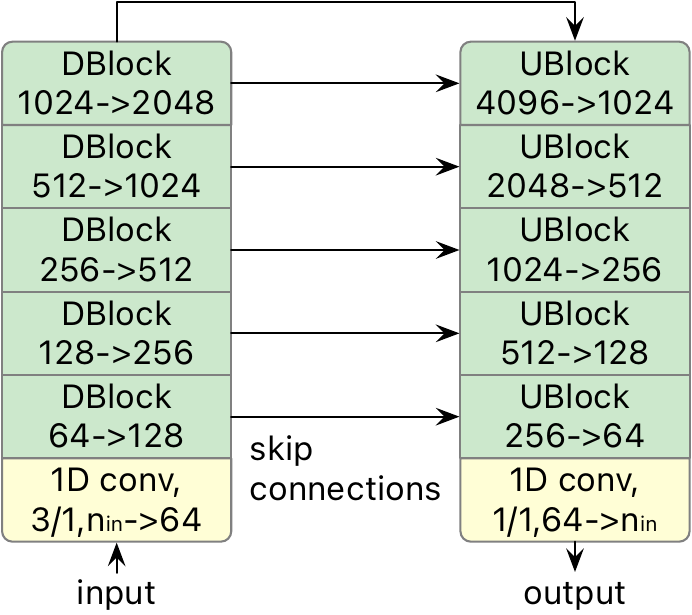}
}
\caption{
The $1D$ convolutional diffusion model used.
}
\label{fig:diff_design}
\end{figure}

\noindent \textit{Diffusion Model (Diff.)}: 
Diffusion models~\cite{ho2020denoising} learn the data distribution by simulating the diffusion of data points through the latent space by  gradually eliminating Gaussian noise within a Markov chain. 
Our design is inspired by the U-Net~\cite{ronneberger2015u}, and employs a $1D$ convolutional network as illustrated in Figure~\ref{fig:diff_design}.

Similar to our GAN and VAE implementations, the diffusion model also utilizes downsample blocks (see Figure~\ref{fig:gan_design}.a) and upsample blocks (see Figure~\ref{fig:gan_design}.b). 
Differing from the GAN and VAE implementations, we incorporated a skip connection akin to~\cite{ronneberger2015u}. 
The input to each upsample block is concatenated with the intermediate representation transmitted via the skip connection, which aid the network in capturing localized features of the data~\cite{ronneberger2015u}. 

\section{Experiment}
We introduce the dataset, experiment setup, and the experiment results in this section. 
During all training stages, the optimization algorithm executes for 400 epochs using the AdamW optimizer~\cite{loshchilov2017decoupled} with a batch size of 64.
Our learning rate scheduler follows the 1cycle learning rate policy~\cite{smithsuperconvergence2017}. 
All code, data, full experiment results, details, and hyperparameters are available on our companion website~\cite{companionsite}. 

\subsection{Dataset.}\label{sec:experiment_dataset}
We use the UCR Archive~\cite{dau2019ucr} and the UEA Archive~\cite{bagnall2018uea} in our experiments.\footnote{The University of East Anglia (UEA) has contributed several datasets to the UCR Archive, and this archive is often referred to as the ``UCR/UEA Archive". We refer to the UCR/UEA Archive as the ``UCR Archive" to avoid confusion in our discussion.} 
The UCR Archive contains 128 univariate classification datasets, and the UEA Archive contains 30 multivariate classification datasets.
For each of the datasets in each archive, we randomly extract the following data splits: pretraining (50\%), training (30\%), validation (10\%), test (10\%). 
All splits are mutually exclusive, and the pretraining set does not contain labels.
\subsection{Experiment Setup.}
We perform a four-stage experimental pipeline for each dataset: 1) Pretraining, 2) Fine-Tuning, 3) Validation, and 4) Testing.
In Pretraining, we utilize three distinct experimental setups, each corresponding to a different method type.
For methods that do not require pretraining, Stage 1 is skipped and the allocated pretraining set is ignored.
For methods that do not use generated time series, we use the pretraining set to pretrain the model.
For methods involving a data generator, if the data generator needs to be trained, we use the pretraining set to train the generative model. 
We use the generator to generate $n_{\text{gen}}$ time series to pretrain the model with.

We want to leverage the fact an unlimited quantity of data can be generated when using data generators, but remain mindful of computational costs.
We consider the average number of time series in each dataset of the source archive when determining $n_{\text{gen}}$. For datasets from the UCR Archive, this threshold is 1494, and for the UEA Archive dataset it is 3398. 
If the size of the pretraining set is below this threshold, we generate the threshold number of time series. Otherwise, we generate the same number of time series in the pretraining set. 

\subsection{Experiment Results.}
The top-10 methods are presented in Table~\ref{tab:top_ten}, which consolidates the results presented in Table~\ref{tab:summary_ucr}. Table~\ref{tab:summary_ucr} compares the average rank of each combination of network, PTM, and data generator on the UCR and UEA Archives. 
A method is a specific combination of a backbone model, PTM, and data generator.
For each dataset in an Archive, we rank each method according to TSC performance. 
The average rank of a method on the Archive is the mean of its rankings on each of its datasets.

We include three baselines: experiments with no pretraining, and 1NN with Euclidean Distance (ED) and Dynamic Time Warping distance (DTW). 
The 1NN baselines are considered simple yet effective for time series classification problems~\cite{rakthanmanon2012searching, bagnall2017great,dau2019ucr}.
A total of 60 combinations and baselines are compared.
\textit{NG} column indicates where ``no generator" was used: in cases of pretraining, the pretraining set was used, and where no pretraining was conducted the PTM is ``N/A". 
\begin{table}[!ht]
    \centering
    \caption{The 10 best methods on the UCR Archive and UEA Archive classification tasks.}
    \resizebox{0.99\columnwidth}{!}{
    \begin{tabular}{l||l|l}
        & \textbf{UCR Archive} & \textbf{UEA Archive} \\ \hline \hline
        1 & ResNet+TS2Vec+GAN &  ResNet+MixingUp+MG \\
        2 & ResNet+MixingUp+MG &  ResNet+MixingUp+GAN \\
        3 & ResNet+TS2Vec+SW &  ResNet+MixingUp+Diff \\
        4 & ResNet+TS2Vec+$\beta$-VAE &  ResNet+MixingUp+NG \\
        5 & ResNet+TimeCLR+MG &  ResNet+TimeCLR+GAN \\
        6 & ResNet+MixingUp+GAN &  ResNet+TS2Vec+Diff \\
        7 & ResNet+TimeCLR+Diff. &  ResNet+TS2Vec+$\beta$-VAE \\
        8 & ResNet+TimeCLR+$\beta$-VAE &  Transformer+TS2Vec+Diff \\
        9 & ResNet+MixingUp+RW &  ResNet+TimeCLR+SW \\
        10 & ResNet+TimeCLR+GAN &  ResNet+TimeCLR+RW \\ 
    \end{tabular}}
    \label{tab:top_ten}
    \vspace{-1em}
\end{table}

\begin{table*}[!t]
    \centering
    \caption{The values are the average ranks for each configuration of experiments, testing combinations of the backbone models, PTMs, and data generators. Ranks are computed for each archive separately, and are highlighted by row, comparing data generators using the same backbone and PTM: \textbf{bold} values correspond to the top ranking method and \underline{underlined} values correspond to the runner-up.
    }
    \footnotesize
    \begin{tabular}{l|l|l||c|ccc|ccc}
        \textbf{Archive} & \textbf{Backbone} & \textbf{PTM} & \textbf{NG} & \textbf{RW} & \textbf{SW} & \textbf{MG} & \textbf{GAN} & \textbf{$\beta$-VAE} & \textbf{Diff.} \\ \hline \hline
        \multirow{12}{*}{UCR} & 1NN ED & N/A & 40.41 & - & - & - & - & - & - \\ 
        & 1NN DTW & N/A & 32.84 & - & - & - & - & - & - \\ \cline{2-10}
        & \multirow{5}{*}{ResNet} & N/A & 24.16 & - & - & - & - & - & - \\
        & ~ & TimeCLR & 34.56 & 23.90 & 23.27 & \textbf{22.21} & 22.73 & \underline{22.38} & \underline{22.38} \\ 
        & ~ & TS2Vec & 23.44 & 23.23 & \underline{21.95} & 23.23 & \textbf{21.71} & 22.04 & 22.94 \\
        & ~ & MixingUp & 37.89 & 22.61 & 23.47 & \textbf{21.82} & \underline{22.30} & 22.88 & 23.72 \\ 
        & ~ & TF-C & 42.03 & 42.12 & \textbf{40.30} & 42.77 & \underline{41.05} & 41.57 & 42.31 \\ \cline{2-10}
        & \multirow{5}{*}{Transformer} & N/A & 32.87 & - & - & - & - & - & - \\
        & ~ & TimeCLR & 32.89 & 29.91 & 28.60 & 28.16 & \textbf{27.30} & 28.58 & \underline{28.09} \\ 
        & ~ & TS2Vec & \underline{27.59} & 29.17 & 29.70 & 28.50 & 29.66 & \textbf{26.45} & 28.27 \\ 
        & ~ & MixingUp & 29.75 & 29.54 & 28.41 & \underline{28.17} & 29.41 & 30.35 & \textbf{26.73} \\ 
        & ~ & TF-C & 42.38 & \underline{42.03} & 43.21 & 42.23 & 43.80 & 42.88 & \textbf{41.16} \\  
        \hline \hline
        \multirow{12}{*}{UEA} & 1NN ED & N/A & 43.87 & - & - & - & - & - & - \\ 
        & 1NN DTW & N/A & 36.42 & - & - & - & - & - & - \\ \cline{2-10}
        & \multirow{5}{*}{ResNet} & N/A & 30.13 & - & - & - & - & - & - \\
        & ~ & TimeCLR & 27.38 & 25.60 & \underline{25.57} & 28.15 & \textbf{25.08} & 28.73 & 30.87 \\ 
        & ~ & TS2Vec & 26.75 & 26.73 & 26.00 & 28.82 & 32.32 & \underline{25.42} & \textbf{25.27} \\
        & ~ & MixingUp & 24.92 & 29.25 & 27.07 & \textbf{22.53} & \underline{24.53} & 27.53 & 24.75 \\ 
        & ~ & TF-C & \textbf{38.92} & 41.43 & 45.12 & 43.10 & \underline{41.02} & 44.93 & 41.57 \\ \cline{2-10}
        & \multirow{5}{*}{Transformer} & N/A & 26.92 & - & - & - & - & - & - \\ 
        & ~ & TimeCLR & 29.43 & \textbf{26.22} & \underline{27.58} & 29.68 & 28.30 & 28.72 & 27.67 \\ 
        & ~ & TS2Vec & \underline{27.57} & 27.63 & 29.58 & 29.43 & 28.30 & 29.97 & \textbf{25.43} \\ 
        & ~ & MixingUp & 30.35 & 26.95 & 29.12 & 26.90 & \underline{26.72} & 26.98 & \textbf{26.03} \\ 
        & ~ & TF-C & 33.88 & 34.13 & \underline{32.75} & 38.25 & \textbf{32.52} & 34.67 & 38.55 \\  
    \end{tabular}
    \label{tab:summary_ucr}
\end{table*}
We observe the following from the top-10 table:
\begin{enumerate}
    \item The majority of methods in Table~\ref{tab:top_ten} are models pretrained with generated time series. 
    This indicates data generators generally improve pretrained models.
    \item The TimeCLR, TS2Vec, and MixingUp PTMs have demonstrated superior performance compared to TF-C. 
    TimeCLR appears seven times in the table, TS2Vec six times, and MixingUp seven times. 
    \item The three advanced generative models (GAN, $\beta$-VAE, and Diffusion model) appear 12 times between the rankings of both archives. 
    The three simpler models (RW, SW, and MG) appear seven times. 
    This suggests generators with a more adept capability at capturing the data distribution have a slight advantage over simpler methods. 
    However, the difference in average rank performance between these two groups of data generation methods is marginal. 
    
    \hspace{\enumerateparindent}It might be surprising that randomized methods like RW and SW perform similarly to other methods. 
    However, because the contrastive nature of the PTMs, valid positive pairs can be generated even from random time series; TimeCLR can form a valid positive pair from random walk by applying diverse data augmentation techniques to the same original time series.
    Furthermore, as demonstrated in~\cite{yeh2023toward}, it is possible to transfer knowledge across different data domains using pretraining methods.
    \item We observe in Table~\ref{tab:summary_ucr} that, in 10 out of 16 cases, pretraining models with real data may have led to degradation in performance.
    This may indicate the negative impact stems from data scarcity, and may be alleviated using generative models.
    \item When considering architectures, ResNet surpasses the Transformer. Most of the top-10 methods utilize the ResNet backbone. 
\end{enumerate}

\begin{figure}[ht]
\centerline{
\includegraphics[width=0.99\linewidth]{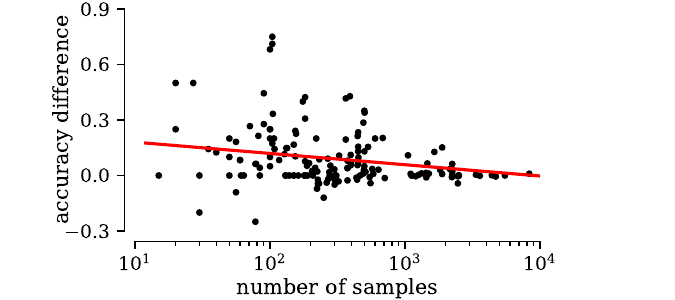}
}
\caption{
The $x$-axis is the size of allocated pretraining set and the $y$-axis the difference in accuracy between ResNet+MixingUp+MG and ResNet+MixingUp+NG.
Positive values indicate where the MG variant outperforms the NG variant.
The fitted line (\textcolor{red}{red}) shows the inverse correlation between dataset size and the performance gain with the data generator. 
Each point is a dataset from either the UCR or UEA Archive.
}
\label{fig:acc_size}
\end{figure}

To ascertain if the time series generators effectively address the data scarcity issue, we  examine the correlation between the size of the pretraining set and the relative accuracy between  methods pretrained with generated data and those pretrained with real data.
In Figure~\ref{fig:acc_size}, the accuracy difference between ResNet+MixingUp+MG and ResNet+MixingUp+NG; these methods were compared because ResNet+MixingUp+MG ranks highly\footnote{The method ranks second in the UCR Archive and first in the UEA Archive.}. 
We include a fitted line to emphasize the correlation between the $x$ and $y$-axes.

The overall trend indicates the use of a data generator is more beneficial when the dataset is small (e.g., fewer than 1000 samples). 
As the pretraining set size increases, the benefits are less pronounced. 
Thus, the adoption of data generators indeed helps alleviate the issue of data scarcity.



\section{Conclusion}
We explore improving the performance of PTMs by utilizing time series generative models. 
By generating vast quantities of time series data for pretraining, we can improve the classification accuracy on the both UCR Archive and UEA Archive by addressing the data scarcity issue. 
We consider different types of generative models, and have included a systematic study of combining different data generators and pretrained methods.
Future work includes improvements to pretraining through superior generative models, an ensemble of such models, and the feasibility of constructing a universal time series generator for pretraining, which encapsulates some commonality in  distributions across time series domains.

\bibliographystyle{ACM-Reference-Format}
\bibliography{section/reference_short.bib}

\end{document}